\documentclass{article}
\usepackage[T1]{fontenc}
\usepackage{iclr2027_conference,times}
\usepackage{graphicx}
\usepackage{xcolor}

\usepackage{amsmath,amsfonts,bm}

\def\eqref#1{equation~\ref{#1}}

\def\1{\bm{1}}

\def\vzero{{\bm{0}}}

\def\vb{{\bm{b}}}
\def\vc{{\bm{c}}}

\def\ve{{\bm{e}}}

\def\vm{{\bm{m}}}

\def\vv{{\bm{v}}}

\def\vx{{\bm{x}}}

\def\vz{{\bm{z}}}

\def\mI{{\bm{I}}}

\DeclareMathAlphabet{\mathsfit}{\encodingdefault}{\sfdefault}{m}{sl}
\SetMathAlphabet{\mathsfit}{bold}{\encodingdefault}{\sfdefault}{bx}{n}

\newcommand{\E}{\mathbb{E}}
\newcommand{\Ls}{\mathcal{L}}
\newcommand{\R}{\mathbb{R}}

\renewcommand{\eqref}[1]{\textup{(\ref{#1})}}

\usepackage{amssymb}
\usepackage{amsthm}
\usepackage{mathtools}
\usepackage{booktabs}
\usepackage{array}
\usepackage{multirow}
\usepackage{subcaption}
\usepackage{flafter}
\usepackage{float}
\usepackage{placeins}
\usepackage{microtype}
\usepackage{hyperref}
\usepackage{url}
\usepackage{cleveref}

\definecolor{kitblue}{RGB}{17,39,74}
\definecolor{kitorange}{RGB}{255,51,0}
\hypersetup{
  colorlinks=false,
  pdfborder={0 0 1},
  citebordercolor={0 1 0},
  linkbordercolor={1 0 0},
  urlbordercolor={0 1 1}
}
\crefname{assumption}{Assumption}{Assumptions}
\Crefname{assumption}{Assumption}{Assumptions}
\crefname{hypothesis}{Hypothesis}{Hypotheses}
\Crefname{hypothesis}{Hypothesis}{Hypotheses}
\crefname{proposition}{Proposition}{Propositions}
\Crefname{proposition}{Proposition}{Propositions}

\newcommand{\method}{\textsc{KiT}}

\newcommand{\vepsilon}{\boldsymbol{\epsilon}}

\theoremstyle{definition}

\theoremstyle{remark}

\title{\method{}: A Foundation Model for Financial \\ Time-Series Forecasting using Diffusion \\ Transformers}

\iclrfinalcopy

\newcommand{\appref}[1]{Appendix~\ref*{#1}}

\newcommand{\maineqref}[1]{Eq.~(\ref*{#1}) of the main paper}

\begin{document}
\maketitle

\begin{center}
    \vspace{-1.8cm}
    \large \textbf{Boyu Zhang$^{1,*}$, Haorui Li$^{2,*}$} \\
    \vspace{0.5em}
    \normalsize
    $^1$University of California, Los Angeles \quad $^2$Southeast University \\
    \vspace{0.5em}
    \texttt{bobo8496@ucla.edu},  \texttt{haoruileee@gmail.com}
    \vspace{0.8em}
\end{center}

\begingroup
\renewcommand{\thefootnote}{\fnsymbol{footnote}}
\footnotetext[1]{Contributed equally.}
\endgroup

\begin{abstract}

Financial candlestick forecasting is fundamental to quantitative investment, yet it remains exceptionally challenging due to extremely low signal-to-noise ratios and vast heterogeneity across markets and instruments. Existing approaches have largely attempted to introduce deep learning to capture hidden temporal features, but most adopt an auto-regressive formulation, which leads to error accumulation during inference. Meanwhile, general-purpose time-series foundation models are not tailored to the unique structure of k-line data and yield unsatisfactory performance on downstream candlestick forecasting tasks. To tackle these problems, we introduce \textbf{KiT}, a \textbf{K}-line D\textbf{i}ffusion \textbf{T}ransformer foundation model, and reformulate future prediction as conditional path generation via flow matching: given a historical context window, the model generates an ensemble of plausible future OHLCV trajectories. We pre-train KiT at multiple parameter scales on billions of candlestick bars spanning multiple markets and timescales. Across three markets and seven resolutions, KiT attains a mean return RankIC of $0.057$ and a mean volatility RankIC of $0.66$, leading at every timescale and outperforming both task-specific financial forecasters and general time-series foundation models. Code will be available at:    \url{https://github.com/Luciferbobo/KiT}.

\end{abstract}

\section{Introduction}
\label{sec:introduction}

A candlestick bar aggregates open, high, low, close, and volume (OHLCV) over a fixed interval, commonly called K-lines, is the atomic unit in which financial markets are recorded, disseminated and acted upon. Forecasting them underpins nearly every stage of the quantitative investment pipeline, from alpha discovery and portfolio construction to risk control and execution scheduling. Yet K-line forecasting remains one of the least forgiving regimes in time-series modeling, for three primary reasons. First, the signal-to-noise ratio is extremely low~\cite{fama1970efficient,lopezdeprado2018advances}. Second, the data are profoundly heterogeneous. Absolute price levels span three orders of magnitude across instruments, volatility regimes differ across markets, and a single instrument behaves differently at one-minute and daily granularity. Third, and most often neglected, the quantity being predicted is intrinsically a distribution rather than a value: the tail risk, uncertainty width and path dependence that a downstream allocation or risk decision actually consumes cannot be expressed by a point forecast at all.

Existing approaches address at most one of these challenges. Early
deep-learning work applies LSTMs and GRUs~\cite{fischer2018deep,zhang2017stock}
and convolutional architectures that treat candlestick charts as
images~\cite{chen2022encoding} to capture temporal dependencies, but these
models are trained per market and horizon with limited transferability. More
recent Transformer forecasters such as iTransformer~\cite{liu2024itransformer},
PatchTST~\cite{nie2023patchtst}, and Temporal Fusion
Transformers~\cite{lim2021temporal} achieve stronger results on general
benchmarks; TFT also predicts quantiles to quantify uncertainty.
General-purpose time-series foundation
models~\cite{ansari2024chronos,das2024timesfm,woo2024moirai,rasul2024lagllama,liu2024timer,goswami2024moment} bring pre-training at scale, and Chronos normalizes values before tokenization.
The recent domain-specific
model Kronos~\cite{shi2026kronos} closes part of this gap by pre-training on
a large multi-market K-line corpus, but inherits two limitations: it
discretizes each bar through a learned quantizer~\cite{bsq}, making the
mapping lossy and leaving the decoder free to emit geometrically impossible
candles; and it decodes autoregressively, leading to error accumulation during inference. Error grows larger as the inference sequence extends. Meanwhile, non-autoregressive generative models have emerged as a promising alternative. Among diffusion-based methods, pioneering works such as TimeGrad~\cite{rasul2021timegrad} and CSDI~\cite{tashiro2021csdi} introduce conditional forecasting and imputation, while subsequent models like TSDiff~\cite{kollovieh2024tsdiff}, Diffusion-TS~\cite{yuan2024diffusionts}, and the non-stationary NsDiff~\cite{fan2025nsdiff} have achieved strong results on standard benchmarks. Furthermore, flow-matching variants such as FlowTS~\cite{hu2025flowts} and TSFlow~\cite{kollovieh2025tsflow} replace stochastic diffusion with deterministic ODE paths to enable faster sampling. In the financial domain specifically, DiffSTOCK~\cite{koa2024diffstock} has explored denoising diffusion for stock prediction. However, all of these generative methods have been developed and evaluated at benchmark scale on homogeneous, single-domain data, and none have been demonstrated as foundation models for financial markets.

Our starting point is that the three difficulties above are best attacked
jointly, by changing \emph{what} is generated, \emph{how} it is represented,
and \emph{where} conditioning enters. We reformulate forecasting as
conditional path generation: given a history window, the model
generates the entire future horizon in one shot as a sample from a
flow matching~\cite{liu2023rectifiedflow,sd3} transport, rather than one bar at a time.
So there is no
temporal rollout and hence no error accumulation by construction. To confront heterogeneity, we recode every bar into a five-dimensional scale-free and exactly invertible state built from log ratios: overnight gap, body, upper and lower shadow, and volume. This represents price movements in relative terms, allowing a \$5 stock and a \$2,000 stock to share a common scale. We impose a conditioning discipline dictated by the data: signals that are constant over a window (market, sector, instrument, time scale, flow time) modulate every layer, whereas signals that vary bar by bar (clock, session position, calendar, trading events) are added directly to token embeddings, where their time alignment survives. History and horizon share one token stream, with history left un-noised, so attention is bidirectional over the full window and the conditioning context is preserved exactly.

We instantiate this design as \textbf{KiT}
(\textbf{K}-line D\textbf{i}ffusion \textbf{T}ransformer). It adopts the diffusion transformer \cite{peebles2023dit} architecture as the backbone, trained via flow matching \cite{lipman2023flowmatching}. We train a family at three
parameter scales ($31$M, $101$M, $284$M) under an identical data protocol, on
billions of bars spanning U.S.\ equities, Chinese A-shares and
cryptocurrencies at seven granularities from one minute to one day, with
leakage-free normalization statistics fitted only on the training period.
Classifier-free guidance~\cite{cfg} over both instrument identity and the
history itself provides an inference-time knob trading sample diversity against
fidelity to context. Across markets and instruments, KiT attains
state-of-the-art RankIC and IC, substantially outperforming both
task-specific financial forecasters and general-purpose time-series foundation
models.

\smallskip
\noindent Our main contributions are:
\begin{itemize}
    \item We propose KiT, the first diffusion-based K-line foundation model that reformulates candlestick forecasting as conditional path generation via flow matching, eliminating the error accumulation inherent in auto-regressive approaches. 
    \item We design a domain-specific candlestick anatomy that is invertible to OHLCV, paired with robust normalization and rich multi-source conditioning, enabling a single diffusion model to generalize across markets, instruments, and timescales. This design eliminates the need for an additional tokenizer, thereby avoiding tokenization errors. 
    \item We pre-train KiT at three parameter scales on billions of bars from diverse markets and show that KiT achieves state-of-the-art results on comprehensive metrics, substantially surpassing existing baselines across various temporal scales. Source code and pre-trained weights will be available.

\end{itemize}
\section{Related Work}
\label{sec:background-related}
\label{sec:conjecture}
\label{sec:related}

\paragraph{Time Series Forecasting.}
Time series forecasting is a fundamental problem in machine learning with broad applications.
In financial markets, accurate forecasting of price dynamics is particularly challenging due to low signal-to-noise ratios, non-stationarity, and heavy-tailed distributions~\citep{cont2001empirical}.
Classical statistical methods, including ARIMA~\citep{box1970time}, GARCH~\citep{bollerslev1986generalized} for volatility modeling, and exponential smoothing~\citep{hyndman2008forecasting}, remain common baselines, while machine-learning approaches such as gradient-boosted trees (XGBoost~\citep{chen2016xgboost}, LightGBM~\citep{ke2017lightgbm}) and random forests offer improved flexibility.
However, these methods struggle with high-dimensional, non-stationary financial data and cannot capture complex cross-asset dependencies.
Transformer-based models have since become the dominant paradigm: Informer~\citep{zhou2021informer}, Autoformer~\citep{wu2021autoformer}, and FEDformer~\citep{zhou2022fedformer} address long-range attention efficiency, while PatchTST~\citep{nie2023patchtst}, TimesNet~\citep{wu2023timesnet}, and Crossformer~\citep{zhang2023crossformer} introduce patching, multi-period structure, and cross-dimension modeling.
Notably, \citet{zeng2023dlinear} show that a single linear layer can match many Transformer variants, prompting channel-oriented designs such as iTransformer~\citep{liu2024itransformer} and multi-scale mixing in TimeMixer~\citep{wang2024timemixer}.
Recent works further advance the field: FreDF~\citep{wang2025fredf} mitigates label correlation bias, ICTSP~\citep{lu2025ictsp} casts forecasting as in-context prediction, and \citet{lau2025fastandslow} introduce dual-stream online adaptation.
Despite this progress, all these methods target point forecasting of continuous series and do not natively handle the structured geometry of candlestick data, where high, low, open and close satisfy rigid ordering constraints~\citep{huang2024structuralohlc}, nor produce full distributional forecasts capturing tail risk and path dependence.

\paragraph{Time Series Foundation Models.}
The success of large language models has motivated time-series foundation models (TSFMs) trained on massive multi-domain corpora for zero- or few-shot transfer.
Representative examples include Chronos~\citep{ansari2024chronos}, TimesFM~\citep{das2024timesfm}, Moirai~\citep{woo2024moirai}, MOMENT~\citep{goswami2024moment}, Timer~\citep{liu2024timer}, Lag-Llama~\citep{rasul2024lagllama}, Time-LLM~\citep{jin2024timellm}, Sundial~\citep{liu2025sundial}, FinCast~\citep{zhu2025fincast}, and LENS~\citep{xu2025lens}.
These models support transfer and, in models such as Chronos, normalized value representations.
In the financial domain, Kronos~\citep{shi2026kronos} is the first foundation model for K-line data, tokenizing OHLCV via binary spherical quantization and pre-training autoregressively on 12\,billion bars; however, discrete tokenization loses continuous price geometry and autoregressive decoding accumulates error over the horizon~\citep{arora2022exposure}.
Complementary financial LLMs (BloombergGPT~\citep{wu2023bloomberggpt}, FinGPT~\citep{yang2023fingpt}) address NLP tasks but do not generate numerical price paths, and market simulators like MarS~\citep{li2024mars} operate at the order level rather than the candlestick level.
\method{} is also a foundation model for financial time series but differs fundamentally: it operates in continuous space via flow matching rather than discrete tokens, and generates the entire forecast horizon non-autoregressively in a single pass.

\paragraph{Diffusion Models for Time Series.}
Denoising diffusion models~\citep{ho2020ddpm,song2021scorebased,song2021ddim} and their ODE-based successor, Flow Matching~\citep{lipman2023flowmatching,liu2023rectifiedflow}, form the modern generative backbone; architecturally, DiT~\citep{peebles2023dit} shows that Transformer-based diffusion scales effectively~\citep{happyhorse26,seedance2026}.
For time series, early works such as TimeGrad~\citep{rasul2021timegrad}, CSDI~\citep{tashiro2021csdi}, D$^3$VAE~\citep{li2022d3vae}, and TimeDiff~\citep{shen2023timediff} demonstrate diffusion-based forecasting and imputation. Further advances include TSDiff~\citep{kollovieh2024tsdiff}, TMDM~\citep{li2024tmdm}, mr-Diff~\citep{shen2024mrdiff}, and Diffusion-TS~\citep{yuan2024diffusionts}.
In finance, FTS-Diffusion~\citep{huang2024ftsdiffusion} addresses irregularity and scale-invariance but operates on daily aggregates rather than multi-resolution candlesticks.
Controllable financial generation is also studied~\citep{tanaka2025cofindiff,zhang2026tfcodit}.
TSFlow~\citep{kollovieh2025tsflow} pairs flow matching with Gaussian process priors, representing the closest flow-matching baseline; TimeDiT~\citep{cao2025timedit} combines DiT with diffusion for general time-series tasks but lacks domain-specific candlestick structure.
Recent works such as LLaDA~\citep{nie2025llada}, MDLM~\citep{sahoo2024mdlm}, and ELF~\citep{hu2026elf} demonstrate that diffusion models have the potential to rival or replace autoregressive generation.
No prior work applies diffusion foundation model to financial candlestick generation, so \method{} fills this gap.

\begin{figure}[t]
  \centering
  \includegraphics[width=\textwidth]{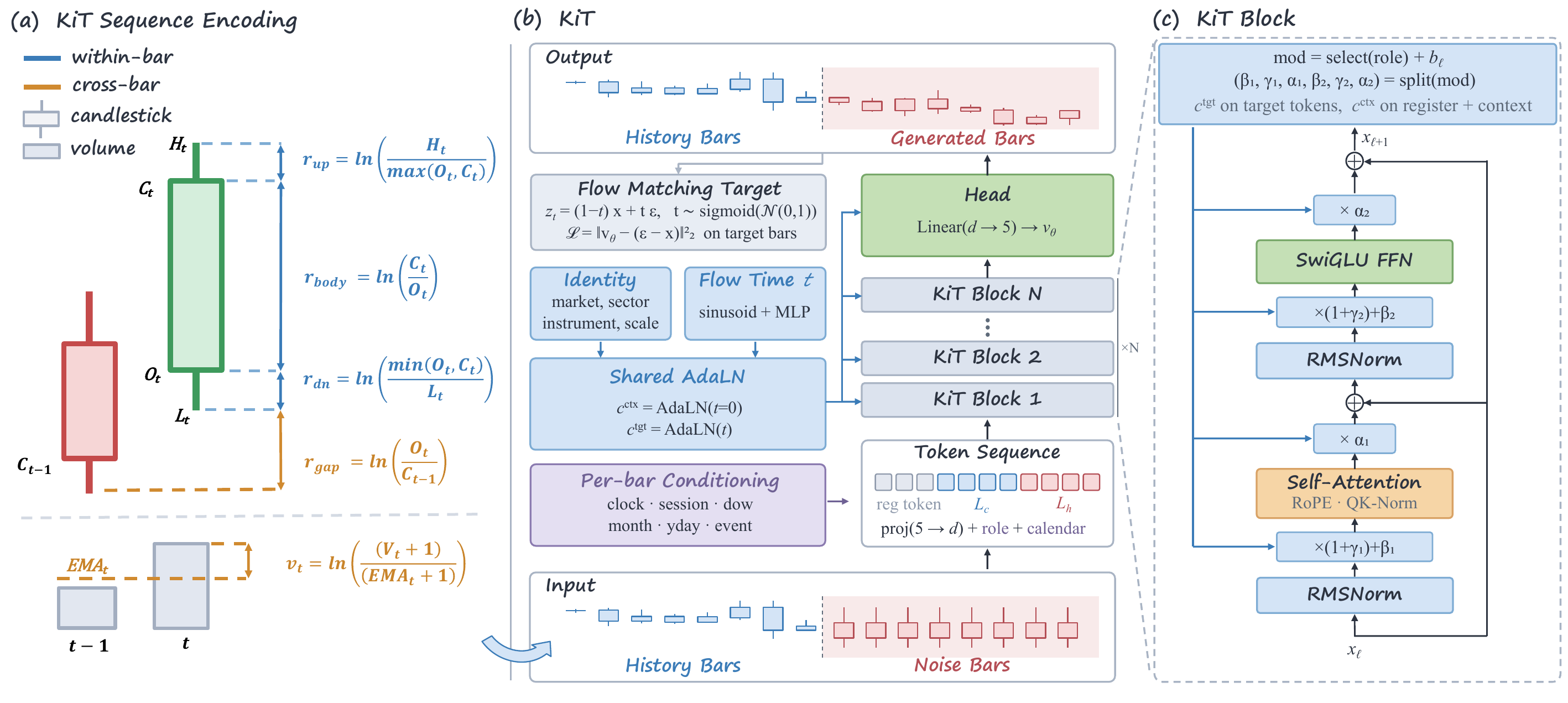}
  \caption{\textbf{The structure of \method{}.}
  \textbf{(a)} Each bar of raw OHLCV is encoded as a five-dimensional log-ratio state $x_t=(r_{\mathrm{gap}}, r_{\mathrm{body}}, r_{\mathrm{up}}, r_{\mathrm{dn}}, v_t)$, which is the state the diffusion model operates on.
  \textbf{(b)} \method{} places the history and forecast spans into one token sequence. The history is returned bit-identical and only the forecast span is filled in with generated bars.
  \textbf{(c)} The \method{} block employs QK-Norm and SwiGLU to improve training stability. Signals that are constant over the window modulate every layer through a shared AdaLN trunk, whereas signals that vary per bar are added directly to the token embeddings.}
  \label{fig:headline}
\end{figure}

\section{KiT}
\label{sec:structured-path-generation}
\label{sec:method}

We present \method{}, a foundation model that casts multi-horizon
candlestick forecasting as conditional path generation via flow matching.  The overall pipeline is illustrated in
Figure~\ref{fig:headline}: raw OHLCV bars are encoded into a
scale-free five-dimensional state~\textbf{(a)}, history and horizon
are assembled into a single token sequence and processed by the
\method{} backbone~\textbf{(b)}, whose repeating block is detailed
in~\textbf{(c)}.  Below we first review the flow matching
preliminaries (\S\ref{sec:rf}), then describe each component in turn.

\subsection{Preliminaries}
\label{sec:rf}

Flow Matching~\citep{liu2023rectifiedflow,sd3} defines a
deterministic transport between data $\vx_0 \sim p_{\text{data}}$ and
noise $\vepsilon \sim \mathcal{N}(\vzero,\mI)$ along a linear
interpolation path.  The noised state at flow time
$t\!\in\![0,1]$ is
\begin{equation}\label{eq:forward}
  \vz_t \;=\; (1-t)\,\vx_0 \;+\; t\,\vepsilon,
\end{equation}
so the ground-truth velocity field is
$\vv^{*} = \vepsilon - \vx_0$.  A neural network
$\vv_\theta(\vz_t,t)$ is trained to predict this velocity by
minimising the mean-squared error
\begin{equation}\label{eq:loss-rf}
  \Ls_{\text{RF}}
  \;=\;
  \E_{t,\,\vx_0,\,\vepsilon}\!\Big[
    \bigl\lVert \vv_\theta(\vz_t,\,t) - (\vepsilon - \vx_0)
    \bigr\rVert^{2}
  \Big],
\end{equation}
where $t$ is drawn from a logit-normal distribution,
$t = \sigma\!\bigl(\mathcal{N}(0,1)\bigr)$, to concentrate training
on intermediate noise levels~\citep{sd3}.  At inference, samples are
obtained by integrating the learned ODE
$\mathrm{d}\vz_t = \vv_\theta(\vz_t,t)\,\mathrm{d}t$ from
$\vz_1 = \vepsilon$ back to $\vz_0 \approx \vx_0$ with an Euler
solver.

\subsection{KiT Sequence Encoding}
\label{sec:repr}

Raw OHLCV values are ill-suited as diffusion states.  Absolute prices
span three orders of magnitude across instruments, volatility regimes differ across time scales, and
volume units are market-specific.  Normalising to zero mean does not
remove these heterogeneities: two instruments with the same
standardised open can still have wildly different price-to-shadow
ratios.

We therefore decompose each candlestick bar into a five-dimensional
log-ratio state $\vx_t \in \R^5$
(Figure~\ref{fig:headline}\textbf{(a)}):
\begin{equation}\label{eq:candle5d}
  \vx_t
  \;=\;
  \begin{pmatrix}
    r_{\mathrm{gap}} \\[2pt]
    r_{\mathrm{body}} \\[2pt]
    r_{\mathrm{up}} \\[2pt]
    r_{\mathrm{dn}} \\[2pt]
    v
  \end{pmatrix}_{\!t}
  \;=\;
  \begin{pmatrix}
    \ln(O_t / C_{t-1}) \\[2pt]
    \ln(C_t / O_t) \\[2pt]
    \ln\bigl(H_t / \max(O_t,C_t)\bigr) \\[2pt]
    \ln\bigl(\min(O_t,C_t) / L_t\bigr) \\[2pt]
    \ln\bigl((V_t+1)/(E_t+1)\bigr)
  \end{pmatrix},
\end{equation}
where $E_t$ is a strictly causal exponential moving average of volume.
This encoding has three key properties.
(i)~Invertibility.  Given the previous close
$C_{t-1}$, the five coordinates recover $(O,H,L,C,V)_t$ via a
closed-form chain of exponentials, so no information is lost.
(ii)~Scale-freeness.  Because every coordinate is a log
ratio, the representation is invariant to the absolute price level and
to multiplicative rescaling of volume, allowing a single model to span
instruments and markets.
(iii)~Structural legality.  The upper and lower shadow
coordinates satisfy $r_{\mathrm{up}} \ge 0$ and
$r_{\mathrm{dn}} \ge 0$ by construction.  At generation time a simple
clamp enforces this constraint, guaranteeing that every decoded candle
satisfies $H_t \ge \max(O_t,C_t)$ and $L_t \le \min(O_t,C_t)$.

Before entering the model, each coordinate is normalised with
per-(market, timescale, feature) robust statistics: we apply median absolute deviation (MAD) scaling followed by a $\tanh$ soft-clip to
bound the values, ensuring stable training without discarding outliers.

\subsection{\method{} Architecture}
\label{sec:arch}

\method{} arranges the input as a single flat sequence
(Figure~\ref{fig:headline}\textbf{(b)}):
\begin{equation}\label{eq:seq}
  [\;\underbrace{r_1,\ldots,r_{n_r}}_{\text{register}}
  \;\mid\;
  \underbrace{x_1^{\mathrm{ctx}},\ldots,x_{L_c}^{\mathrm{ctx}}}_{\text{context}}
  \;\mid\;
  \underbrace{z_1^{\mathrm{tgt}},\ldots,z_{L_h}^{\mathrm{tgt}}}_{\text{target}}\;].
\end{equation}
The $n_r$ register tokens are learnable parameters that serve as
global attention sinks and information aggregators.
The context segment contains the $L_c$ clean (un-noised) historical
bars; the target segment contains the $L_h$ future bars, which carry
noise $\vz_t$ during training and pure Gaussian noise at inference.
Each five-dimensional bar is projected to the model dimension $d$ by a
linear layer, to which a \emph{role embedding}
$\ve_{\mathrm{role}} \in \{\ve^{\mathrm{ctx}},\ve^{\mathrm{tgt}}\}$
and a set of \emph{calendar embeddings} are added element-wise.
History is never noised and is returned bit-identical at the output, so
context information is preserved exactly.

\method{} receives two categories of conditioning signals, handled by
distinct mechanisms matched to their granularity. \emph{Window-constant (identity) signals} such as market, sector,
instrument and timescale identifiers are each mapped to learned
embeddings and summed into a single base conditioning vector
$\vc_{\mathrm{base}} \in \R^d$.  Following the PixArt-$\alpha$ shared
AdaLN design~\citep{chen2024pixart}, one shallow MLP maps the sum of
$\vc_{\mathrm{base}}$ and a sinusoidal flow-time embedding $\ve_t$ to a
set of modulation parameters that are broadcast to \emph{every}
transformer layer.  Crucially, this trunk is evaluated \emph{twice}:
once with the actual flow time $t$ to produce modulations
$\vc^{\mathrm{tgt}}$ applied to target tokens, and once with $t{=}0$
to produce $\vc^{\mathrm{ctx}}$ applied to context tokens.
Setting $t{=}0$ for context is natural: the historical bars carry no
noise, so their modulation should correspond to the clean-data end of
the flow.  This dual evaluation adds negligible cost, since the trunk
MLP is shared and lightweight, yet provides each token role with
appropriately calibrated normalisation statistics. \emph{Per-bar (calendar) signals}, including Fourier-encoded intraday
time, session position, day-of-week, event flags, month, and
year-of-day, vary across bars and are known deterministically at
inference.  These are embedded and added directly to the token
representation before the first transformer layer, preserving their
temporal alignment without inflating the AdaLN parameter count.

After the final transformer layer, only the $L_h$ target-position
hidden states are extracted.  These are passed through an RMSNorm,
modulated by a final set of AdaLN parameters from $\vc^{\mathrm{tgt}}$,
and projected back to five dimensions by a linear layer, yielding the
velocity prediction $\hat{\vv} \in \R^{L_h \times 5}$.

\subsection{\method{} Block}
\label{sec:block}

Each of the $N$ transformer layers follows the \method{} block design
shown in Figure~\ref{fig:headline}\textbf{(c)}, which extends the
standard DiT block~\citep{peebles2023dit} with role-aware dual modulation.

At every layer $l$, a token's modulation vector is selected by its role
and offset by a per-layer learnable bias:
\begin{equation}\label{eq:adaln-select}
  \vm_l
  \;=\;
  \mathrm{select}\!\bigl(\mathrm{role},\;
    \vc^{\mathrm{ctx}},\;\vc^{\mathrm{tgt}}\bigr)
  \;+\; \vb_l,
\end{equation}
where $\vb_l$ is initialised to zero.  The vector $\vm_l$ is split
into six groups of modulation parameters
$(\beta_1,\gamma_1,\alpha_1,\beta_2,\gamma_2,\alpha_2)$, each in
$\R^d$.

The pre-attention path applies RMSNorm~\citep{zhang2019rmsnorm} with affine modulation:
\begin{equation}\label{eq:attn-branch}
  \vx \;\leftarrow\; \vx
    \;+\;
    \alpha_1 \odot
    \operatorname{Attn}\!\bigl(
      \operatorname{RMSNorm}(\vx) \odot (1+\gamma_1) + \beta_1
    \bigr),
\end{equation}
where $\operatorname{Attn}$ denotes bidirectional multi-head
self-attention with Rotary Position Embeddings
(RoPE)~\citep{su2024roformer} and QK-Norm~\citep{henry2020querykey}
applied to the queries and keys for training stability.  Attention is
\emph{not} causal: every token, including future ones, attends to the
full sequence, which lets the model exploit inter-bar dependencies
within the noisy horizon.

The post-attention path mirrors the structure:
\begin{equation}\label{eq:ffn-branch}
  \vx \;\leftarrow\; \vx
    \;+\;
    \alpha_2 \odot
    \operatorname{SwiGLU}\!\bigl(
      \operatorname{RMSNorm}(\vx) \odot (1+\gamma_2) + \beta_2
    \bigr),
\end{equation}
using a SwiGLU~\citep{shazeer2020glu} feed-forward network instead of
the standard MLP.

All output projections of the AdaLN trunk that produce the gate
scalars $(\alpha_1,\alpha_2)$, as well as the per-layer biases
$\vb_l$, are initialised to zero.  Consequently, at the start of
training every block acts as an identity, giving
$\vv_\theta(\cdot) = \vzero$ regardless of the input.  This ensures
the initial velocity prediction is zero, so the model begins from a
well-defined starting point and avoids large, uninformative gradient
updates early in training~\citep{peebles2023dit}.

\subsection{Training and Inference}
\label{sec:training}

\paragraph{Training objective.}
The loss is the masked MSE of Equation~\eqref{eq:loss-rf}, evaluated
\emph{only} over the $L_h$ target positions:
\begin{equation}\label{eq:loss-kit}
  \Ls
  \;=\;
  \E_{t,\,\vx_0,\,\vepsilon}\!\Bigg[
    \frac{1}{L_h}
    \sum_{i=1}^{L_h}
    \bigl\lVert
      \vv_\theta(\vz_t,\,t)_i - (\vepsilon_i - \vx_{0,i})
    \bigr\rVert^{2}
  \Bigg],
\end{equation}
where context tokens contribute to attention but are excluded from the
loss.  The flow time $t$ is sampled via the logit-normal schedule.

\paragraph{Classifier-free guidance for KiT.}
We design a classifier-free guidance (CFG) mechanism tailored for candlestick chart prediction to adjust the predictive distribution~\citep{cfg}. Specifically, we use three stochastic condition-dropout schemes during training: the instrument and sector embeddings are replaced by a shared $\texttt{[UNK]}$ token; all identity signals (market, sector, instrument, timescale) are replaced by a $\texttt{[NULL]}$ vector, and the entire history context is zeroed out.

\paragraph{Inference.}
At test time, starting from $\vz_1 = \vepsilon \sim
\mathcal{N}(\vzero,\mI)$, we integrate the learned ODE with an Euler
solver.  To exploit both
conditioning axes, we apply dual classifier-free
guidance:
\begin{equation}\label{eq:dual-cfg}
  \hat{\vv}
  \;=\;
  \vv_\theta
  \;+\;
  (w_{\mathrm{id}}-1)\bigl(\vv_\theta - \vv_{\varnothing\mathrm{id}}\bigr)
  \;+\;
  (w_{\mathrm{hist}}-1)\bigl(\vv_\theta - \vv_{\varnothing\mathrm{hist}}\bigr),
\end{equation}
where $\vv_{\varnothing\mathrm{id}}$ and
$\vv_{\varnothing\mathrm{hist}}$ are the velocity predictions under
the identity-dropped and history-dropped conditions, respectively, and
$w_{\mathrm{id}},w_{\mathrm{hist}} \ge 1$ control the guidance
strength along each axis.  Drawing $M$ independent samples and
decoding each through the invertible map of
Equation~\eqref{eq:candle5d} yields an ensemble of $M$ plausible
OHLCV trajectories, from which any distributional statistic can be computed directly.
\section{Experiments}
\label{sec:evaluation}

To evaluate the effectiveness of KiT, we collect a decade of candlestick data from multiple markets and build three models with different parameter sizes (\S\ref{sec:experiment-setup}). 
We assess return forecasting and volatility prediction performance (\S\ref{sec:forecasting}), and also conduct backtests (\S\ref{sec:backtest}). Then we conduct ablation studies to validate the design choices of each component in KiT.(\S\ref{sec:abst} and \appref{app:ablation}).

\newlength{\ricSignWd}
\newlength{\ricMantWd}
\newlength{\ricNameWd}

\newcommand{\ricp}[1]{%
  \makebox[\ricSignWd][r]{\phantom{$-$}}%
  \makebox[\ricMantWd][l]{#1}%
}
\newcommand{\ricn}[1]{%
  \makebox[\ricSignWd][r]{$-$}%
  \makebox[\ricMantWd][l]{#1}%
}
\newcommand{\ricb}[1]{%
  \makebox[\ricSignWd][r]{\phantom{$-$}}%
  \makebox[\ricMantWd][l]{\textbf{#1}}%
}
\newcommand{\ricsetup}{%
  \tiny\setlength{\tabcolsep}{3.4pt}%
  \settowidth{\ricSignWd}{$-$}%
  \settowidth{\ricMantWd}{\textbf{.0000}}%
  \settowidth{\ricNameWd}{Historical return bootstrap}%
}

\begin{table*}[t]
\centering
\caption{\textbf{Return forecasting by resolution.} Evaluated across three markets in the test window. Each entry is the RankIC between the predicted mean terminal return and the realized close-to-close log-return sum.}
\label{tab:hf-val-ret}
\label{tab:hf-val-ic}
\begingroup\ricsetup
\resizebox{\textwidth}{!}{%
\begin{tabular}{w{l}{\ricNameWd}rrrrrrrr}
\toprule
Configuration & 1m $\uparrow$ & 5m $\uparrow$ & 15m $\uparrow$ & 30m $\uparrow$ & 1h $\uparrow$ & 2h $\uparrow$ & 1d $\uparrow$ & Mean $\uparrow$ \\
\midrule
KiT & \ricb{.0215} & \ricb{.0576} & \ricb{.0526} & \ricb{.0474} & \ricb{.0383} & \ricb{.0651} & \ricb{.1168} & \ricb{.0571} \\
\midrule
Kronos-base & \ricp{.0158} & \ricp{.0426} & \ricp{.0442} & \ricn{.0064} & \ricn{.0128} & \ricp{.0514} & \ricp{.1046} & \ricp{.0342} \\
Kronos-base-FT & \ricp{.0192} & \ricp{.0536} & \ricp{.0484} & \ricp{.0408} & \ricp{.0304} & \ricp{.0622} & \ricp{.1101} & \ricp{.0521} \\
Sundial & \ricp{.0148} & \ricp{.0386} & \ricp{.0434} & \ricn{.0082} & \ricn{.0046} & \ricn{.0214} & \ricp{.0837} & \ricp{.0209} \\
Chronos-Bolt extended & \ricn{.0362} & \ricp{.0088} & \ricp{.0186} & \ricn{.0964} & \ricn{.0068} & \ricn{.2159} & \ricp{.0052} & \ricn{.0461} \\
Chronos-2 & \ricp{.0168} & \ricp{.0286} & \ricp{.0362} & \ricn{.0184} & \ricn{.0162} & \ricn{.0648} & \ricp{.0556} & \ricp{.0054} \\
TimesFM 2.5 & \ricp{.0126} & \ricp{.0284} & \ricn{.0160} & \ricn{.0797} & \ricn{.0089} & \ricn{.0747} & \ricp{.0116} & \ricn{.0181} \\
TimesFM 3 & \ricp{.0046} & \ricp{.0338} & \ricp{.0374} & \ricn{.0168} & \ricp{.0294} & \ricn{.0586} & \ricp{.0479} & \ricp{.0111} \\
\midrule
Historical return bootstrap & \ricn{.0296} & \ricn{.0128} & \ricn{.0346} & \ricp{.0267} & \ricp{.0341} & \ricp{.0020} & \ricp{.0982} & \ricp{.0120} \\
\midrule
TimeGrad adapted & \ricp{.0136} & \ricp{.0088} & \ricp{.0362} & \ricp{.0046} & \ricn{.0248} & \ricp{.0574} & \ricp{.0792} & \ricp{.0250} \\
Diffusion-TS adapted & \ricp{.0090} & \ricp{.0284} & \ricp{.0206} & \ricp{.0292} & \ricn{.0746} & \ricp{.0198} & \ricn{.1654} & \ricn{.0190} \\
TSFlow adapted & \ricp{.0174} & \ricp{.0554} & \ricp{.0466} & \ricp{.0228} & \ricp{.0102} & \ricp{.0596} & \ricp{.0673} & \ricp{.0399} \\
\bottomrule
\end{tabular}}
\endgroup
\vspace{-0.2cm}
\end{table*}

\begin{table*}[t]
\centering
\caption{\textbf{Volatility prediction by resolution.} Evaluated across three markets in the test window. Each entry is the RankIC between the predicted mean realized volatility and the realized sample volatility.}
\label{tab:hf-val-vol}
\begingroup\ricsetup
\resizebox{\textwidth}{!}{%
\begin{tabular}{w{l}{\ricNameWd}rrrrrrrr}
\toprule
Configuration & 1m $\uparrow$ & 5m $\uparrow$ & 15m $\uparrow$ & 30m $\uparrow$ & 1h $\uparrow$ & 2h $\uparrow$ & 1d $\uparrow$ & Mean $\uparrow$ \\
\midrule
KiT & \ricb{.7599} & \ricb{.6947} & \ricb{.6900} & \ricb{.5919} & \ricb{.5958} & \ricb{.6652} & \ricb{.6225} & \ricb{.6600} \\
\midrule
Kronos-base & \ricp{.5900} & \ricp{.4934} & \ricp{.3955} & \ricp{.3049} & \ricp{.4699} & \ricp{.5281} & \ricp{.4498} & \ricp{.4617} \\
Kronos-base-FT & \ricp{.6193} & \ricp{.5478} & \ricp{.4466} & \ricp{.3671} & \ricp{.4974} & \ricp{.5306} & \ricp{.4727} & \ricp{.4974} \\
Sundial & \ricp{.6024} & \ricp{.5477} & \ricp{.5164} & \ricp{.4591} & \ricp{.5691} & \ricp{.5534} & \ricp{.4486} & \ricp{.5281} \\
Chronos-Bolt extended & \ricp{.5766} & \ricp{.4821} & \ricp{.4113} & \ricp{.4289} & \ricp{.3543} & \ricp{.3797} & \ricp{.3310} & \ricp{.4234} \\
Chronos-2 & \ricp{.4427} & \ricp{.3944} & \ricp{.3671} & \ricp{.4092} & \ricp{.4264} & \ricp{.4452} & \ricp{.2260} & \ricp{.3873} \\
TimesFM 2.5 & \ricp{.5184} & \ricp{.4609} & \ricp{.4435} & \ricp{.4537} & \ricp{.5407} & \ricp{.4688} & \ricp{.3096} & \ricp{.4565} \\
TimesFM 3 & \ricp{.3870} & \ricp{.4655} & \ricp{.4117} & \ricp{.3474} & \ricp{.4470} & \ricp{.3977} & \ricp{.2507} & \ricp{.3867} \\
\midrule
GARCH & \ricp{.3520} & \ricp{.5940} & \ricp{.6469} & \ricp{.4850} & \ricp{.5842} & \ricp{.5681} & \ricp{.4255} & \ricp{.5222} \\
\midrule
TimeGrad adapted & \ricp{.0736} & \ricp{.1967} & \ricp{.0056} & \ricn{.1586} & \ricp{.5201} & \ricp{.4418} & \ricp{.1702} & \ricp{.1785} \\
Diffusion-TS adapted & \ricp{.4462} & \ricp{.1787} & \ricp{.1483} & \ricp{.0246} & \ricp{.4236} & \ricp{.3488} & \ricp{.4563} & \ricp{.2895} \\
TSFlow adapted & \ricp{.3244} & \ricp{.3230} & \ricp{.4043} & \ricp{.5658} & \ricp{.5290} & \ricp{.3648} & \ricp{.5250} & \ricp{.4338} \\
\bottomrule
\end{tabular}}
\endgroup
\vspace{-0.2cm}
\end{table*}

\begin{figure*}[t]
  \centering
  \includegraphics[width=\textwidth]{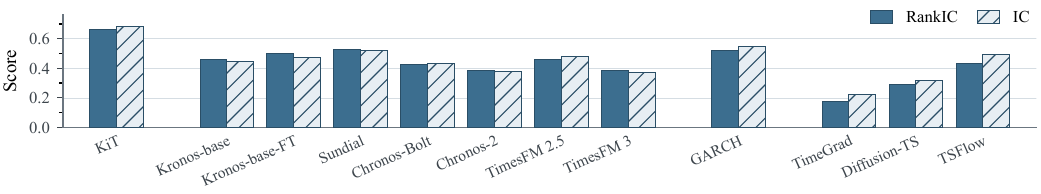}
  \caption{\textbf{Mean RankIC and IC of volatility} on the three-market
  with seven resolutions.}
  \label{fig:vol-ic-rankic}
\vspace{-0.0cm}
\end{figure*}

\begin{figure}[t]
  \centering
  \includegraphics[width=\textwidth]{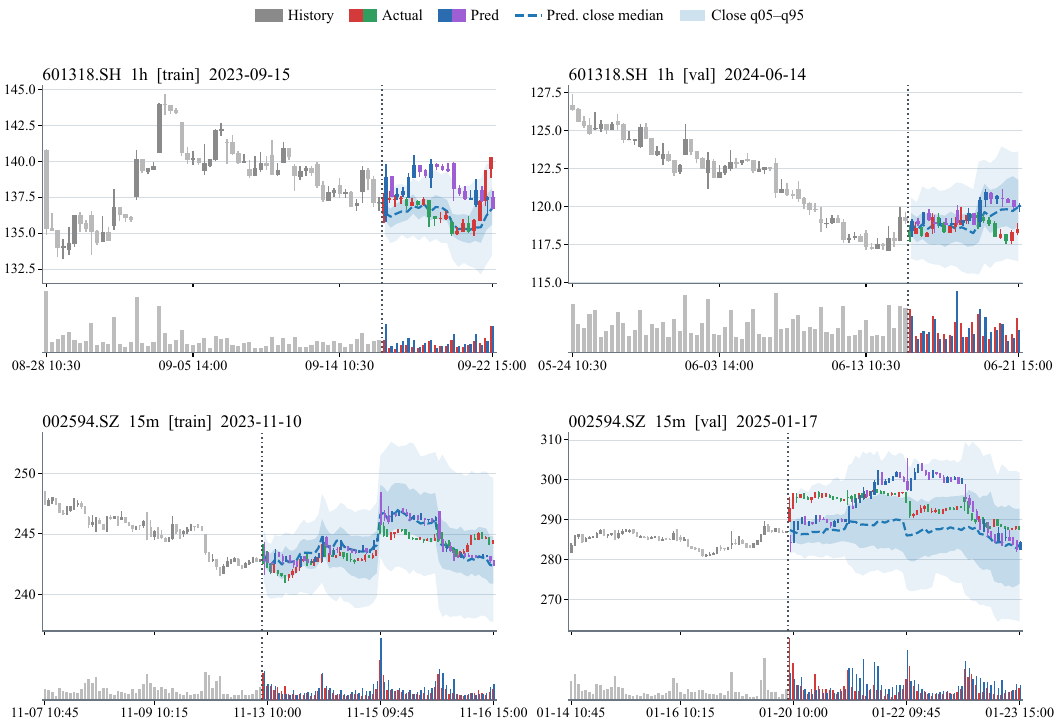}
  \caption{\textbf{Qualitative candlestick forecasts from KiT}. Gray candles are history. Blue and purple represent predictions, while red and green indicate ground truth.}
  \label{fig:kline-cases}
  \vspace{-0.5cm}
\end{figure}

\begin{figure}[t]
  \centering
  \includegraphics[width=\textwidth]{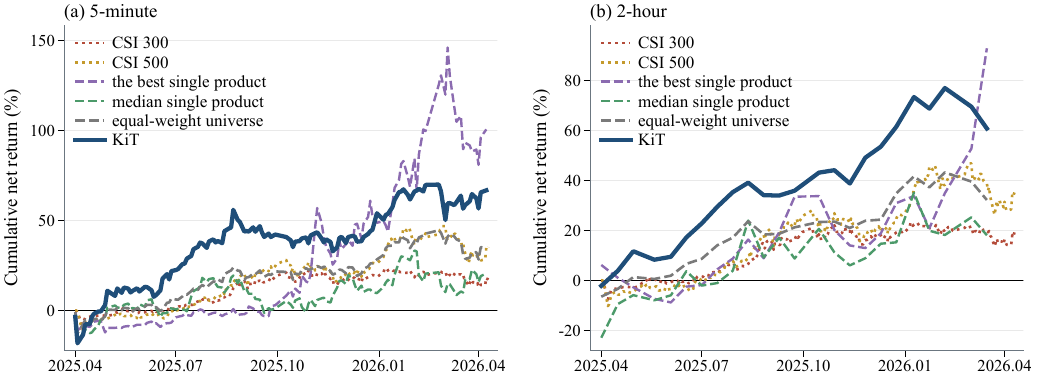}
  \vspace{-0.5cm}
  \caption{\textbf{Return-based long-only backtest} over a one-year period, using KiT to forecast two different timescales.}
  \label{fig:backtest-two}
  \vspace{-0.5cm}
\end{figure}

\subsection{Experimental setup}
\label{sec:experiment-setup}

We collect OHLCV bars from three markets at seven timescales: one, five,
fifteen, and thirty minutes, one and two hours, and one day.
Chinese A-shares cover 3,914 instruments, US equities cover 888 names,
and cryptocurrencies cover 28 products, all from 1 August 2016 to
10 April 2026.
The merged tree contains 4,830 instruments and 3 billion
candlestick bars after aggregating the one-minute source to the coarser
grids. Prices are backward-adjusted. Normalization statistics
(per-market, per-timescale MAD scales) are fitted only on the training
period.

Training uses all bars through 31 December 2023. Validation is 1 January 2024 to 31 December 2025. The test window is 1 January 2026 to 10 April 2026. We train with
AdamW ($\beta_1{=}0.9$, $\beta_2{=}0.95$, weight decay $0.1$), BF16,
global batch 4,096, gradient clipping at $1.0$, and a parameter EMA
with decay $0.9999$. The peak learning rate is $10^{-4}$ after 1,000
warmup updates. Training uses 128 NVIDIA H200 GPUs and continues until
the mean validation loss, averaged equally over the monitored
(market, timescale) buckets, no longer improves. We instantiate three
sizes under a shared data protocol: KiT-S (31M), KiT-M (101M), and
KiT-L (284M). All subsequent evaluations use the largest parameter version, denoted as KiT.

For further details about the model settings, please see \appref{app:pretraining_families}.
At inference we draw 16 samples and average them as the predicted
trajectory. Classifier-free guidance weights default to $1$; we
discuss CFG ablations in \appref{app:cfg}.

We evaluate forecasts with the information coefficient
(IC)~\citep{grinold1989fundamental,lopezdeprado2018advances} and the
rank information coefficient (RankIC)~\citep{spearman1904proof}. IC is the cross-sectional
Pearson correlation of these predictions against outcomes; RankIC is
the corresponding Spearman rank correlation. Higher IC and RankIC are better. We compare KiT against recent state-of-the-art time-series
forecasting methods, including financial foundation models and general
time-series models. For implementation details of the baselines, please see \appref{app:baselines}.

\subsection{Forecasting results}
\label{sec:forecasting}
\label{sec:matched-path-comparison}
\label{sec:price-series}

\paragraph{Return forecasting.}
Table~\ref{tab:hf-val-ret} reports RankIC of predicted terminal return
at seven resolutions on the test window, from 1 January 2026 to 10 April 2026. KiT attains the best mean RankIC of $0.0571$ and the
best mean IC of $0.0491$, leading at every resolution, from one minute
($0.0215$) through one day ($0.1168$). Against Kronos-base-FT, the strongest
financial foundation model on this metric, mean RankIC rises from $0.0521$
to $0.0571$ (about $10\%$ relative) and mean IC from $0.0247$ to $0.0491$.
On return IC the closest baseline is TimeGrad ($0.0446$), a smaller gain of
about $10\%$ relative. TSFlow ($0.0399$) and Kronos-base ($0.0342$) trail
further on RankIC. Per-resolution return IC is in
\appref{app:extended_forecasts}.

\paragraph{Volatility forecasting.}
In Table~\ref{tab:hf-val-vol}, KiT attains the best mean RankIC of
$0.6600$ and the best mean IC of $0.6797$, and leads at every resolution.
Relative to the strongest baselines, this is a gain of $+0.132$ RankIC over
Sundial ($0.5281$, about $25\%$ relative) and $+0.133$ IC over GARCH
($0.5470$, about $24\%$ relative). GARCH ($0.5222$) and Kronos-base-FT
($0.4974$) follow on RankIC. Mean volatility IC is shown in
Figure~\ref{fig:vol-ic-rankic}; per-resolution volatility IC is in
\appref{app:extended_forecasts}.

We include some OHLCV forecast samples to showcase the real-world predictive capabilities of KiT. Gray candles are history; colored candles are the realized
future. The line is the median generated close and the band is a close
quantile interval. As illustrated in Figure~\ref{fig:kline-cases}, KiT can identify key temporal events such as market open and close, at which the predicted results exhibit pronounced price fluctuations and volume surges. Moreover, this capability generalizes consistently across different time scales and products, which is also the behavior the calendar injections are designed to produce. For more forecasting results and price series forecasts, see \appref{app:extended_forecasts}.

\vspace{-0.1cm}
\subsection{Backtest}
\label{sec:backtest}
\vspace{-0.2cm}
To further validate the practical applicability of KiT, we aim to obtain a trading result. Therefore, we rank names by predicted
terminal return and replay a long-only book on A-share
over a one-year period (2025.05-2026.04). Top-$N$ and
rebalance frequency are selected on validation and then frozen. Buys
that would close at a limit-up are skipped. Costs are 12.5~bp on the
buy side and 22.5~bp on the sell side (commission 2.5~bp, stamp 10~bp,
slippage 10~bp per side).

Figure~\ref{fig:backtest-two} shows the backtest results of two timescales after
costs. The KiT book finishes above the equal-weight universe of the
same products and above the median single-name buy-and-hold, also demonstrating excess returns over both CSI 300 and CSI 500. Test
annualized net return is $+65.2\%$ on 5-minute bars (Sharpe $1.70$) and
$+64.4\%$ on 2-hour bars (Sharpe $2.46$).

\vspace{-0.1cm}
\subsection{Ablation Study}
\label{sec:abst}
\vspace{-0.0cm}

\paragraph{Classifier-free guidance.}
We investigate how classifier-free guidance operates on KiT in \appref{app:cfg}. Identity guidance and history guidance move different margins of the same ensemble.
In Equation~\eqref{eq:dual-cfg}, raising $w_{\mathrm{id}}$ pulls samples toward each product's typical volatility, while raising $w_{\mathrm{hist}}$ pulls the terminal return toward the recent window.

\vspace{-0.2cm}
\paragraph{Model scale.}
To assess the scaling capability of KiT, we train different variants of the KiT family: KiT-S, KiT-M, and KiT-L, on the same dataset. The performance of KiT-L has already been presented in above \S\ref{sec:evaluation}. We observe that increasing the parameter count yields modest improvements in both return forecasting and volatility prediction, demonstrating that scaling up the model size enhances overall performance at this data scale~\citep{chickering2026abra,hoffmann2022chinchilla} (\appref{app:pretraining_analysis}).

\vspace{-0.1cm}

We also conduct ablation studies on candlestick representation and condition injection to demonstrate the rationale behind the design of each component in our KiT, see \appref{app:ablation} for details.

\vspace{-0.2cm}
\section{Conclusion}
\label{sec:conclusion}
\label{sec:discussion}
\label{sec:limitations}
\vspace{-0.2cm}

We introduce KiT, a novel framework that models candlestick paths utilizing a flow-matching diffusion transformer. We design a five-dimensional OHLCV representation that strictly preserves candlestick geometry, alongside versatile condition-injection mechanisms tailored to the intrinsic properties of diffusion models. Comprehensive evaluations across multiple markets and time scales at seven distinct resolutions demonstrate that KiT significantly outperforms prior methods in both return and volatility forecasting. Furthermore, theoretical backtests illustrate the strong economic relevance of its generated path forecasts.

Future extensions include integrating higher-dimensional data and fine-tuning for downstream tasks. By obviating the need for a discrete tokenizer, KiT can seamlessly accommodate high-dimensional inputs, such as Level-3 limit order book data and alpha factors. As KiT provides a robust foundation model, exploring targeted fine-tuning remains a promising direction for subsequent research.

\bibliography{iclr2027_conference}
\bibliographystyle{iclr2027_conference}
\clearpage
\appendix
\renewcommand{\topfraction}{.95}
\renewcommand{\bottomfraction}{.90}
\renewcommand{\textfraction}{.05}
\renewcommand{\floatpagefraction}{.75}
\setcounter{topnumber}{5}
\setcounter{bottomnumber}{3}
\setcounter{totalnumber}{7}
\makeatletter
\setlength{\@fptop}{0pt}
\setlength{\@fpsep}{11pt plus 2pt minus 2pt}
\setlength{\@fpbot}{0pt plus 1fil}
\makeatother

\section{Experiment Details}
\label{app:data_protocol}

\subsection{Model families}
\label{app:pretraining_families}

Table~\ref{tab:app_training_models} lists KiT-S/M/L family. All models are trained with
the same seven resolutions. We use different history and prediction window lengths across different K-line time scales, with the specific settings shown in Table \ref{tab:app_windows}. The rationale behind this multi-scale window configuration is to establish an optimal equilibrium between the model computational cost and the intrinsic physical dynamics of financial time series.

\begin{table}[htbp]
\centering
\begin{minipage}[t]{0.42\textwidth}
\centering
\caption{Model configurations.}
\label{tab:app_training_models}
\label{tab:pretraining-series}
\small
\renewcommand{\arraystretch}{1.25} 
\begin{tabular*}{\linewidth}{@{\extracolsep{\fill}}lrrr@{}}
\toprule
 & KiT-S & KiT-M & KiT-L \\
\midrule
Parameters & 31.0M & 101.0M & 283.7M \\
Layers & 8 & 13 & 21 \\
Width & 512 & 768 & 1,024 \\
Heads & 4 & 6 & 8 \\
FFN & 1,408 & 2,048 & 2,816 \\
Steps & 50k & 50k & 50k \\
\bottomrule
\end{tabular*}
\end{minipage}%
\hspace{0.06\textwidth}
\begin{minipage}[t]{0.42\textwidth}
\centering
\caption{Window size.}
\label{tab:app_windows}
\small
\renewcommand{\arraystretch}{1.1} 
\begin{tabular*}{\linewidth}{@{\extracolsep{\fill}}lrrr@{}}
\toprule
Resolution & History & Horizon \\
\midrule
1 minute & 1,200 & 120 \\
5 minutes & 960 & 96 \\
15 minutes & 800 & 64 \\
30 minutes & 640 & 40 \\
1 hour & 400 & 20 \\
2 hours & 360 & 20 \\
1 day & 250 & 20 \\
\bottomrule
\end{tabular*}
\end{minipage}
\end{table}

\subsection{Baseline implementations}
\label{app:baselines}

The settings below describe the close-only baselines. The main-text RankIC tables and Appendix~\ref{app:extended_forecasts} use one comparison. Each method
receives only the observed history and predicts the horizon associated
with that resolution. For a sampled close path, we compute its terminal
close-to-close log-return and the sample standard deviation of its
one-bar log-returns. We average these two summaries over $K$ paths before
computing cross-sectional IC or RankIC. Sundial, the statistical baselines,
and the three adapted generative models use $K{=}16$. The Chronos and
TimesFM interfaces below supply a single point path ($K{=}1$); their
volatility score is the temporal variability of that path. Marginal
quantiles are not treated as independent joint-path samples. All of these
validation runs record inference seed 91000.

\vspace{-0.2cm}
\paragraph{Kronos.}
The official Kronos~\citep{shi2026kronos} release includes a 102M-parameter base checkpoint.
We evaluate that public Kronos-base model with its official close
channel and $K{=}8$ samples. Kronos-base-FT is the same architecture
finetuned on our mixed-market training split; we record the checkpoint
with the best test-set scores.

\vspace{-0.2cm}
\paragraph{Sundial.}
We use the public 128M-parameter \texttt{thuml/sundial-base-128m}
checkpoint~\citep{liu2025sundial} without additional training. The model
takes univariate close prices in their original units, with no future
covariates. Its native generation interface produces 16 future close
paths in FP32, with the requested prediction length set to the evaluation
horizon. The longest horizon here is 120 bars, within the model's native
720-bar prediction limit.

\vspace{-0.2cm}
\paragraph{Chronos-Bolt extended.}
The validation run uses \texttt{amazon/chronos-bolt-small}
(48M parameters)~\citep{ansari2024chronos} in BF16, without finetuning.
It receives the observed close sequence and returns marginal forecasts.
The validation adapter records a single point readout under
\texttt{mean\_quantile\_forecast}. The ``extended'' setting
covers the full requested horizon, including the 96- and 120-bar horizons
that exceed its native 64-bar output block. The resulting sequence is
scored as one point path, not as an ancestral sample ensemble.

\vspace{-0.2cm}
\paragraph{Chronos-2.}
We use the public \texttt{amazon/chronos-2}
checkpoint~\citep{ansari2025chronos2} without financial finetuning.
The close-only adapter forecasts each instrument from its observed
history. The validation run uses BF16 and records one point forecast per case. Its
returned close sequence supplies both the terminal-return and
within-horizon volatility readouts under the shared scoring interface.

\vspace{-0.2cm}
\paragraph{TimesFM 2.5 and TimesFM 3.}
We use \texttt{google/timesfm-2.5-200m-pytorch} and
\texttt{google/timesfm-3.0-pytorch}~\citep{google2025timesfm25,google2026timesfm3}
as public pretrained checkpoints, with observed close prices as the
univariate input and no additional training. The TimesFM 2.5 validation
adapter reads channel 5 of the decoded output as its point forecast.
The TimesFM 3 adapter uses the median ($q_{0.5}$), with the returned mean
as a fallback when a median is unavailable. Both yield one future close
sequence per case. These are point-forecast comparisons; the available
quantile outputs are not converted into joint stochastic trajectories.

\vspace{-0.2cm}
\paragraph{Historical return bootstrap.}
For each case, we form the empirical distribution of one-bar log-returns
from its observed history. Each of 16 forecasts independently draws $H$
returns with replacement from this distribution, where $H$ is the
resolution-specific horizon. Summing the draws gives the terminal
log-return; their sample standard deviation gives the volatility
forecast. This baseline requires no learned parameters. Resampling
individual returns retains the empirical marginal distribution but
does not preserve serial dependence.

\vspace{-0.2cm}
\paragraph{GARCH.}
The validation baseline is a zero-mean Gaussian GARCH(1,1)
process~\citep{bollerslev1986generalized} with fixed coefficients:
\[
  r_{t+1}=\sqrt{v_t}\,\epsilon_{t+1},\qquad
  v_{t+1}=10^{-8}+0.05\,r_{t+1}^{2}+0.90\,v_t,
  \qquad \epsilon_{t+1}\sim\mathcal{N}(0,1).
\]
The variance is initialized from the sample variance of historical
log-returns and filtered through the observed sequence before forecasting.
We simulate 16 paths over the required horizon and use the same return
and volatility summaries as above. The coefficients are fixed across
instruments and resolutions; this run does not fit GARCH parameters
by likelihood maximization.

\vspace{-0.2cm}
\paragraph{Historical training for adapted generative models.}
TimeGrad, Diffusion-TS, and TSFlow are trained separately at each
resolution on historical A-share windows whose targets end by
December 31, 2023; historical validation targets fall in 2024.
Their input history lengths are 512 bars at 1, 5, 15, and 30 minutes,
400 at 1 hour, 360 at 2 hours, and 250 at 1 day. Prediction horizons
match Table~\ref{tab:app_windows}. We express close prices relative to
the last observed close in log space, then map them with an affine
scaler fitted to training-window extrema. The same scaler is reused
at inference and inverted after sampling. Each model receives exactly
10,000 optimizer updates per resolution. ``Terminal-10k'' denotes this
fixed training budget and the final checkpoint; it does not denote
fitting on the 2026 evaluation window or selecting the best evaluation
score.

\vspace{-0.2cm}
\paragraph{TimeGrad adapted.}
We retain the official TimeGrad network~\citep{rasul2021timegrad} and
provide two close-derived channels: normalized log-close and its causal
first difference. Only the generated close channel is inverse-transformed
and scored. The network uses a two-layer LSTM with 40 cells per layer,
dropout 0.1, and 100 diffusion steps per autoregressive future bar.
Training uses batch size 32 and Adam with weight decay $10^{-6}$;
the OneCycle learning-rate schedule peaks at $10^{-2}$.
Inference draws 16 paths from the terminal non-EMA checkpoint.

\vspace{-0.2cm}
\paragraph{Diffusion-TS adapted.}
We use the official close-only Diffusion-TS
model~\citep{yuan2024diffusionts}, with two encoder and two decoder
layers, width 64, four attention heads, and 500 training diffusion
steps under a cosine schedule. Training uses batches of 8 with two
gradient-accumulation steps and an EMA decay of 0.995, updated every
10 optimizer steps. Forecasting uses the terminal EMA weights and
conditional infilling: the normalized observed prefix is supplied with
an explicit mask, while the unknown suffix is initialized to zero.
The native sampler generates 16 paths with 200 sampling steps and
input-gradient refinement (coefficient 0.01, learning rate 0.05).
The 200 sampling steps do not represent a measured count of network
evaluations, since refinement makes additional calls.

\vspace{-0.2cm}
\paragraph{TSFlow adapted.}
We retain TSFlow's official univariate S4 backbone and
Ornstein--Uhlenbeck prior~\citep{kollovieh2025tsflow}, using the
nonseasonal branch without lag features. Its additional scale factor is
the mean absolute normalized history, floored at $10^{-6}$, and is
computed from the observed context only. The prior context uses the
largest complete multiple of $H$ contained in that history. Training
uses batch size 64, Adam at $10^{-3}$, gradient-norm clipping at 0.5,
and the official EMA with decay 0.9999 and an update delay of 128 steps.
The terminal EMA checkpoint generates 16 paths using Euler integration
on 32 time-grid points. We report the grid size rather than relabeling
it as 32 network evaluations.

\begin{figure}[h]
\vspace{-0.0cm}
\centering
\includegraphics[width=\linewidth]{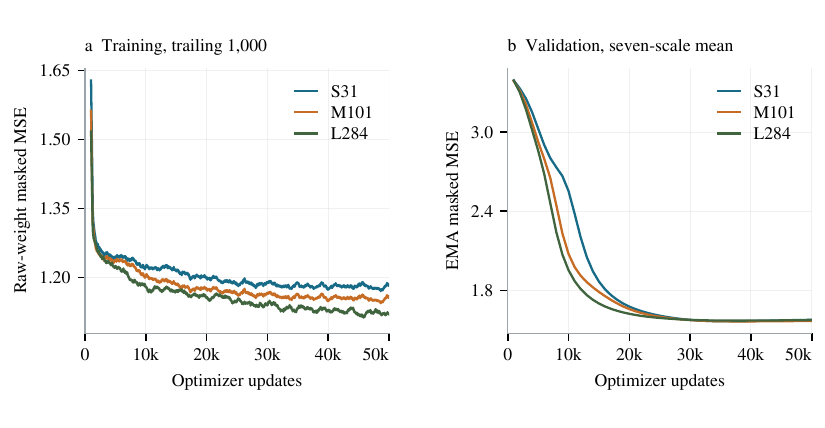}
\vspace{-0.0cm}
\caption{\textbf{Training loss curves and validation loss curves of the model family.} All curves stop at 50k optimizer updates. Training uses a trailing 1,000-update mean; validation averages the seven resolution losses.}
\label{fig:exp-training}
\end{figure}

\begin{figure}[h]
\vspace{-0.0cm}
\centering
\includegraphics[width=0.8\linewidth]{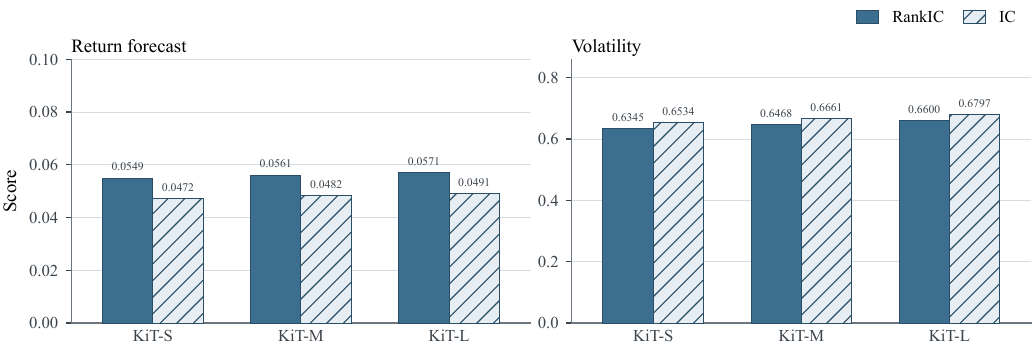}
\vspace{-0.0cm}
\caption{\textbf{Forecast quality across model scale.}}
\label{fig:exp-scale-forecast}
\end{figure}

\vspace{-0.2cm}
\section{Pretraining Scale and Forecast Quality}
\label{app:pretraining_analysis}
\vspace{-0.2cm}

Figure~\ref{fig:exp-training} shows the training and validation loss
curves through 50k optimizer updates for all three model sizes. At 50k,
the trailing 1,000-update training means are 1.183, 1.157, and 1.121 for
the small, medium, and large models. Their equal-resolution validation
means are 1.577, 1.575, and 1.568, respectively. The validation traces
converge to a similar level by about 30k updates.

Figure~\ref{fig:exp-scale-forecast} compares mean return and volatility
scores for KiT-S (31M), KiT-M (101M), and KiT-L (284M) on the same
three-market, seven-resolution test panel. KiT-L is the checkpoint
reported in the main paper, with return RankIC $0.0571$, return IC
$0.0491$, volatility RankIC $0.6600$, and volatility IC $0.6797$.

The RankIC bars follow a mild scaling law~\citep{shi2024scalinglaw}. At the
current data volume, a larger model still improves RankIC~\citep{spearman1904proof}: each step
up in size raises it by about two percent relative, and the
order is KiT-S, then KiT-M, then KiT-L for return and for volatility.
The gain is small, but it has not flattened. Increasing capacity remains
useful on this corpus, in the same direction as the training-loss gap
among the three models.

\section{Extended Forecast Results}
\label{app:extended_forecasts}

We provide here more detailed evaluation results for return and volatility forecasting. The evaluation results below use the same test period as in \S\ref{sec:forecasting}, both being computed in the test window from 1 January 2026 to 10 April 2026.

\begingroup
\setlength{\intextsep}{7pt}

\subsection{Return IC}
\label{app:return_ic_table}

Return IC compares mean predicted and realized terminal close-to-close log-returns.

\makeatletter
\@ifundefined{ricsetup}{%
  \newlength{\ricSignWd}%
  \newlength{\ricMantWd}%
  \newlength{\ricNameWd}%
  \newcommand{\ricp}[1]{%
    \makebox[\ricSignWd][r]{\phantom{$-$}}%
    \makebox[\ricMantWd][l]{#1}%
  }%
  \newcommand{\ricn}[1]{%
    \makebox[\ricSignWd][r]{$-$}%
    \makebox[\ricMantWd][l]{#1}%
  }%
  \newcommand{\ricb}[1]{%
    \makebox[\ricSignWd][r]{\phantom{$-$}}%
    \makebox[\ricMantWd][l]{\textbf{#1}}%
  }%
  \newcommand{\ricsetup}{%
    \tiny\setlength{\tabcolsep}{3.4pt}%
    \settowidth{\ricSignWd}{$-$}%
    \settowidth{\ricMantWd}{\textbf{.0000}}%
    \settowidth{\ricNameWd}{Historical return bootstrap}%
  }%
}{}%
\makeatother

\begin{table*}[h]
\centering
\caption{\textbf{Return IC by resolution.} Higher is better; bold marks the best value in each column.}
\label{tab:app-hf-val-ret-ic}
\begingroup\ricsetup
\resizebox{\textwidth}{!}{%
\begin{tabular}{w{l}{\ricNameWd}rrrrrrrr}
\toprule
Configuration & 1m $\uparrow$ & 5m $\uparrow$ & 15m $\uparrow$ & 30m $\uparrow$ & 1h $\uparrow$ & 2h $\uparrow$ & 1d $\uparrow$ & Mean $\uparrow$ \\
\midrule
KiT & \ricb{.0235} & \ricb{.0552} & \ricb{.0445} & \ricb{.0330} & \ricb{.0241} & \ricb{.0475} & \ricb{.1158} & \ricb{.0491} \\
\midrule
Kronos-base & \ricp{.0072} & \ricp{.0344} & \ricp{.0368} & \ricn{.0244} & \ricn{.0209} & \ricp{.0186} & \ricn{.0048} & \ricp{.0067} \\
Kronos-base-FT & \ricp{.0164} & \ricp{.0506} & \ricp{.0392} & \ricp{.0136} & \ricp{.0125} & \ricp{.0348} & \ricp{.0058} & \ricp{.0247} \\
Sundial & \ricp{.0086} & \ricp{.0074} & \ricp{.0362} & \ricp{.0221} & \ricp{.0152} & \ricp{.0284} & \ricp{.1026} & \ricp{.0315} \\
Chronos-Bolt extended & \ricn{.0364} & \ricp{.0292} & \ricp{.0246} & \ricn{.0724} & \ricp{.0068} & \ricn{.2223} & \ricp{.0276} & \ricn{.0347} \\
Chronos-2 & \ricp{.0188} & \ricp{.0383} & \ricp{.0372} & \ricn{.0402} & \ricn{.0841} & \ricn{.1789} & \ricp{.0962} & \ricn{.0161} \\
TimesFM 2.5 & \ricn{.0124} & \ricp{.0426} & \ricp{.0348} & \ricn{.0286} & \ricp{.0186} & \ricn{.0592} & \ricp{.0924} & \ricp{.0126} \\
TimesFM 3 & \ricn{.0062} & \ricp{.0384} & \ricp{.0352} & \ricn{.0116} & \ricp{.0196} & \ricn{.0242} & \ricp{.1042} & \ricp{.0222} \\
\midrule
Historical return bootstrap & \ricn{.0112} & \ricp{.0199} & \ricn{.0805} & \ricn{.0114} & \ricp{.0194} & \ricn{.0434} & \ricn{.1000} & \ricn{.0296} \\
\midrule
TimeGrad adapted & \ricp{.0212} & \ricp{.0454} & \ricp{.0400} & \ricp{.0314} & \ricp{.0228} & \ricp{.0428} & \ricp{.1086} & \ricp{.0446} \\
Diffusion-TS adapted & \ricn{.0209} & \ricp{.0412} & \ricn{.0096} & \ricp{.0084} & \ricn{.0491} & \ricp{.0386} & \ricn{.0142} & \ricn{.0008} \\
TSFlow adapted & \ricp{.0188} & \ricp{.0412} & \ricp{.0376} & \ricp{.0089} & \ricp{.0216} & \ricp{.0394} & \ricp{.0124} & \ricp{.0257} \\
\bottomrule
\end{tabular}}
\endgroup
\vspace{-0.5cm}
\end{table*}

\vspace{-0.2cm}
\subsection{Volatility IC}
\label{app:vol_ic_table}

Volatility IC compares mean predicted path volatility with realized future volatility.

\begin{table*}[h]
\centering
\caption{\textbf{Volatility IC by resolution.} Higher is better; bold marks the best value in each column.}
\label{tab:app-hf-val-vol-ic}
\begingroup\ricsetup
\resizebox{\textwidth}{!}{%
\begin{tabular}{w{l}{\ricNameWd}rrrrrrrr}
\toprule
Configuration & 1m $\uparrow$ & 5m $\uparrow$ & 15m $\uparrow$ & 30m $\uparrow$ & 1h $\uparrow$ & 2h $\uparrow$ & 1d $\uparrow$ & Mean $\uparrow$ \\
\midrule
KiT & \ricb{.7459} & \ricb{.6749} & \ricb{.6888} & \ricb{.6553} & \ricb{.6621} & \ricb{.7586} & \ricb{.5726} & \ricb{.6797} \\
\midrule
Kronos-base & \ricp{.5654} & \ricp{.4369} & \ricp{.3616} & \ricp{.3175} & \ricp{.4491} & \ricp{.5745} & \ricp{.4377} & \ricp{.4490} \\
Kronos-base-FT & \ricp{.5836} & \ricp{.4695} & \ricp{.3884} & \ricp{.3759} & \ricp{.4667} & \ricp{.5736} & \ricp{.4470} & \ricp{.4721} \\
Sundial & \ricp{.6055} & \ricp{.5178} & \ricp{.5233} & \ricp{.5041} & \ricp{.5757} & \ricp{.5138} & \ricp{.4081} & \ricp{.5212} \\
Chronos-Bolt extended & \ricp{.5830} & \ricp{.4926} & \ricp{.4169} & \ricp{.4907} & \ricp{.4261} & \ricp{.4414} & \ricp{.1562} & \ricp{.4296} \\
Chronos-2 & \ricp{.4156} & \ricp{.3688} & \ricp{.3185} & \ricp{.5068} & \ricp{.3665} & \ricp{.5365} & \ricp{.1272} & \ricp{.3771} \\
TimesFM 2.5 & \ricp{.5172} & \ricp{.4250} & \ricp{.4760} & \ricp{.4784} & \ricp{.5770} & \ricp{.4618} & \ricp{.4369} & \ricp{.4818} \\
TimesFM 3 & \ricp{.3851} & \ricp{.3758} & \ricp{.3810} & \ricp{.3723} & \ricp{.5436} & \ricp{.4191} & \ricp{.1277} & \ricp{.3721} \\
\midrule
GARCH & \ricp{.3810} & \ricp{.5285} & \ricp{.6566} & \ricp{.6001} & \ricp{.5697} & \ricp{.6324} & \ricp{.4610} & \ricp{.5470} \\
\midrule
TimeGrad adapted & \ricp{.1311} & \ricp{.2646} & \ricp{.0467} & \ricn{.1019} & \ricp{.5642} & \ricp{.5463} & \ricp{.1176} & \ricp{.2241} \\
Diffusion-TS adapted & \ricp{.4492} & \ricp{.1991} & \ricp{.1305} & \ricp{.0395} & \ricp{.4833} & \ricp{.4467} & \ricp{.4695} & \ricp{.3168} \\
TSFlow adapted & \ricp{.3817} & \ricp{.4176} & \ricp{.4393} & \ricp{.6422} & \ricp{.5132} & \ricp{.5517} & \ricp{.5289} & \ricp{.4964} \\
\bottomrule
\end{tabular}}
\endgroup
\vspace{-0.2cm}
\end{table*}

\FloatBarrier
\endgroup

\subsection{Price-series forecasting}
\label{app:price_series_sec}
\vspace{-0.3cm}
Additionally, we evaluate the price series forecasting performance in the test window, from January 1 2026 to April 10 2026. The results demonstrate that KiT outperforms both the baseline and fine-tuned Kronos models.

\vspace{-0.2cm}
\begin{table}[h]
\centering
\caption{\textbf{Price-series forecasting.} Values are averaged over the seven resolutions and three markets.}
\label{tab:app-price-series}
\small
\begin{tabular}{lc}
\toprule
Configuration & IC/RankIC $\uparrow$ \\
\midrule
KiT & $\mathbf{0.0372/0.0407}$ \\
\midrule
Kronos-base & $0.0090/0.0148$ \\
Kronos-base-FT & $0.0132/0.0205$ \\
\bottomrule
\end{tabular}
\vspace{-0.3cm}
\end{table}

\subsection{Return forecasting over time}
\label{app:ic_timeseries}

\begin{figure}[p]
\centering
\includegraphics[width=\linewidth]{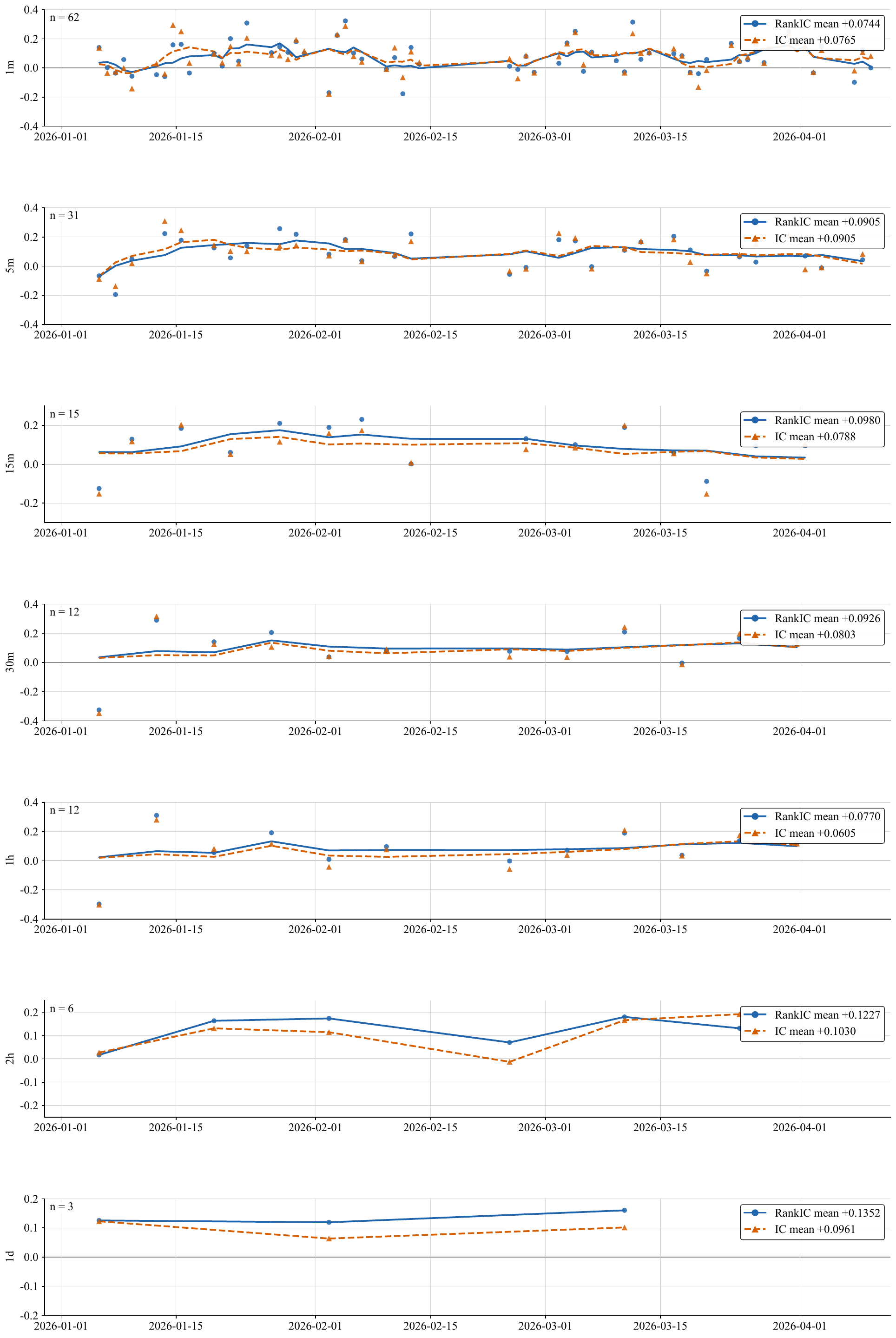}
\caption{A-share return RankIC and IC.}
\label{fig:app-ic-rankic-ashare}
\end{figure}

\begin{figure}[p]
\centering
\includegraphics[width=\linewidth]{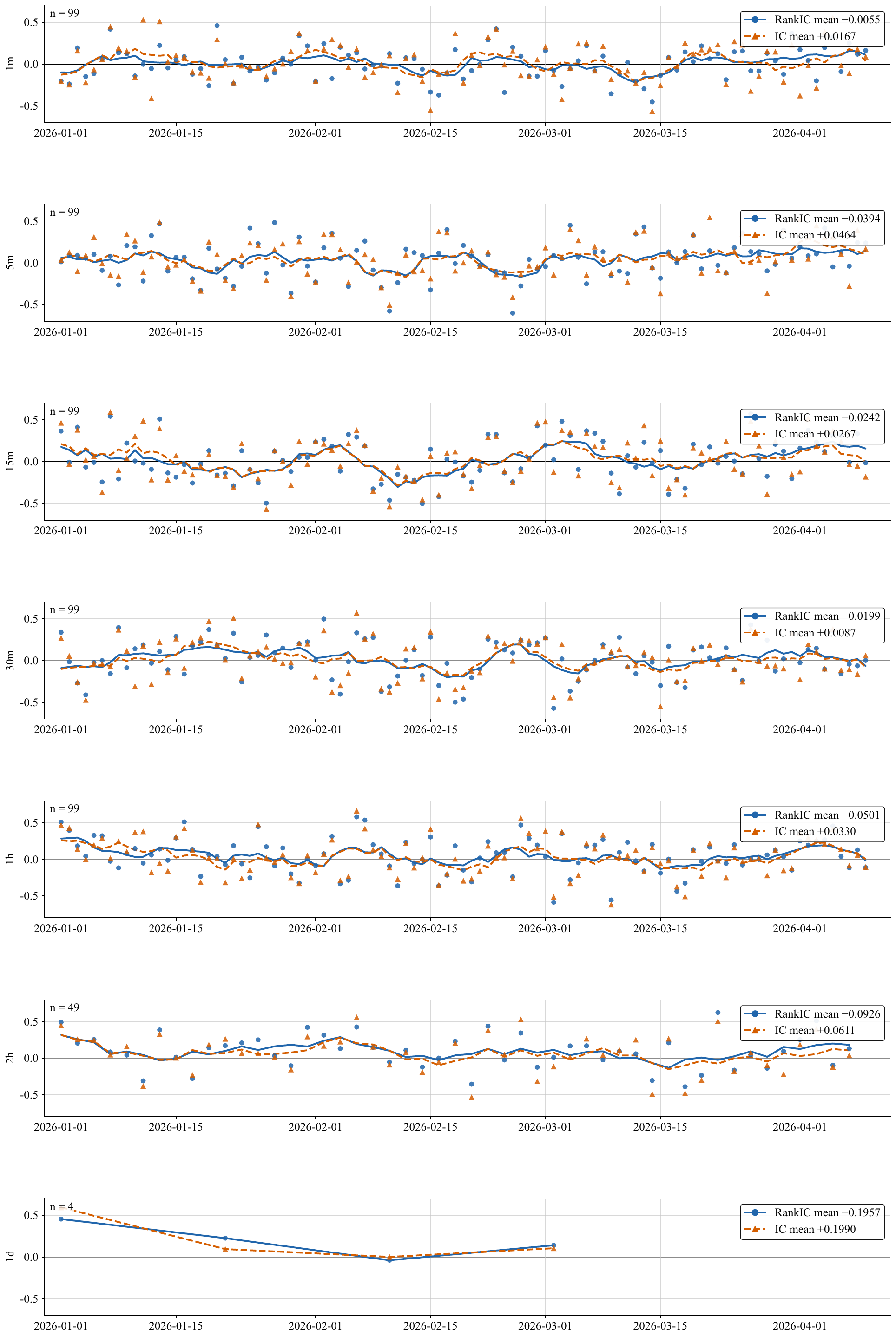}
\caption{Crypto return RankIC and IC.}
\label{fig:app-ic-rankic-crypto}
\end{figure}

\begin{figure}[p]
\centering
\includegraphics[width=\linewidth]{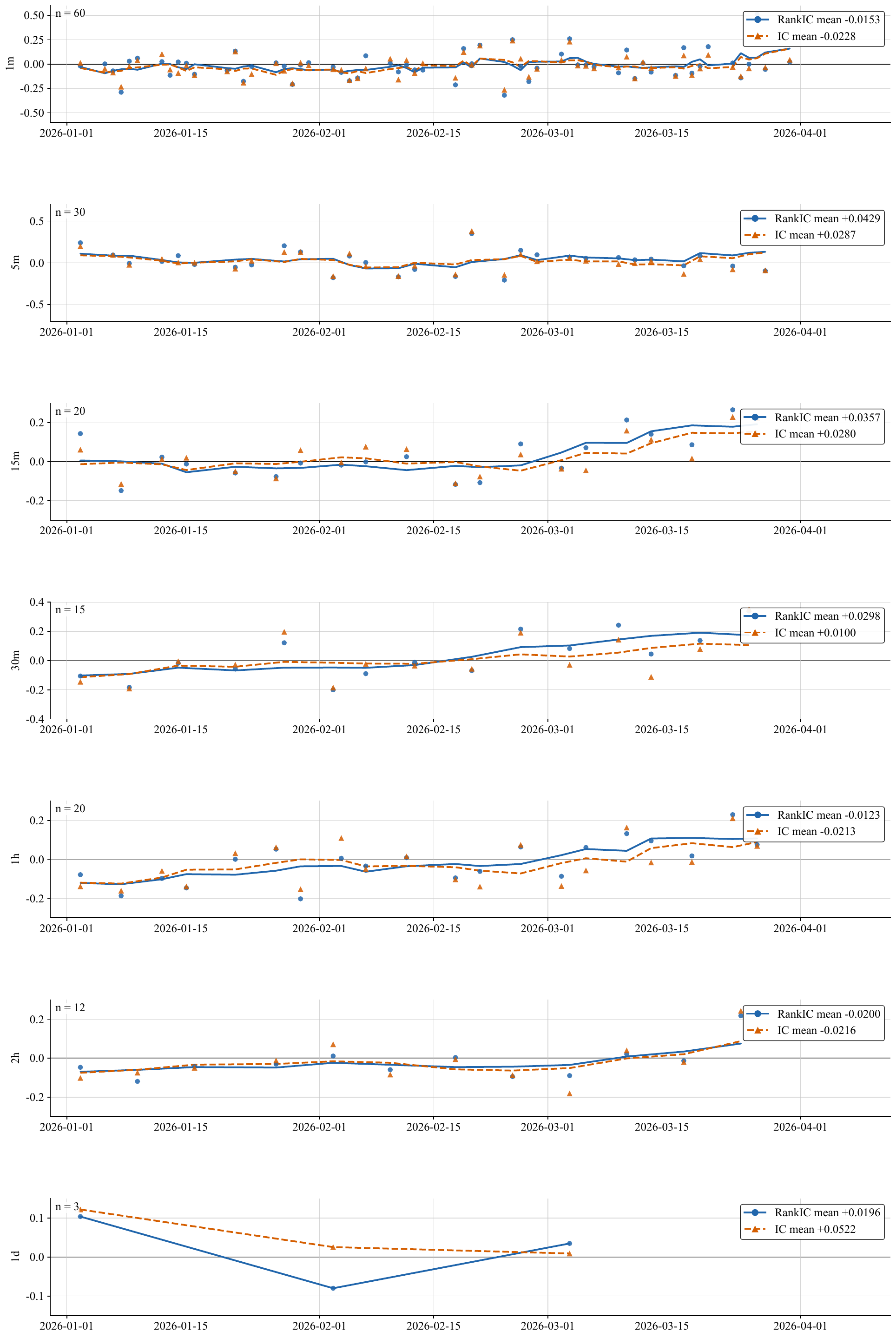}
\caption{US return RankIC and IC.}
\label{fig:app-ic-rankic-us}
\end{figure}

Figures~\ref{fig:app-ic-rankic-ashare}--\ref{fig:app-ic-rankic-us} show how
cross-sectional return RankIC and IC move from one anchor date to the
next, separately for A-shares, crypto, and US equities. Each point is
one anchor; the line is a rolling mean. Evaluation window is from 2026.01.01 to 2026.04.10. Each point here represents a time anchor. So every point in all the figures below represents the average result of tests across all products in each market. For example, a single point in the A-share figure is the average of 3,914 prediction results.

\FloatBarrier
\section{Classifier-free guidance}
\label{app:cfg}

Classifier-free guidance is the usual inference-time control in
conditional generative models~\citep{cfg}. Training randomly replaces
the condition with a null input, so one network represents both the
conditional and the unconditional velocity. At sampling time the two
predictions are extrapolated,
\begin{equation}\label{eq:app-cfg}
  \vv
  \;=\;
  \vv_{\varnothing}
  \;+\;
  w\,(\vv_{c} - \vv_{\varnothing}),
\end{equation}
where $w{=}1$ recovers the conditional model and $w{>}1$ amplifies the
condition. A larger weight makes samples adhere more closely to the
requested condition and reduces their variety. 

KiT applies that tradeoff to a candlestick forecast, whose output is a
distribution over future paths. Two contexts set where the distribution
sits, and \maineqref{eq:dual-cfg} extrapolates them separately.
Identity, market, sector, instrument, and timescale, carries the slow
character of the product: its typical drift, volatility, and candle
geometry. History carries the recent path. Training uses the three
stochastic replacements in the main text. The instrument and sector
embeddings are replaced by a shared $\texttt{[UNK]}$ token; all identity
signals (market, sector, instrument, timescale) are replaced by a
$\texttt{[NULL]}$ vector; and the history context is zeroed. At
inference, $w_{\mathrm{id}}$ scales the identity correction and
$w_{\mathrm{hist}}$ scales the history correction. Raising
$w_{\mathrm{id}}$ moves the ensemble toward the distribution associated
with that product. Raising $w_{\mathrm{hist}}$ moves it toward the
recent window. Either increase concentrates the samples. Both weights
are $1$ in the main evaluation, which is the conditional model.

We select 100 products and, for each, draw $K{=}64$ paths from one
shared initial noise. Each weight is swept over $\{1,2,3,4\}$ while the
other is held at $1$, so differences across a column come only from that
weight. Three diagnostics are averaged over the 100 products.

Prior gap compares the ensemble with the product's own character. For
product $i$, $\sigma_i^{\mathrm{id}}$ is the standard deviation of
log-returns on a long causal window of that instrument, taken before the
forecast anchor and excluding the most recent horizon. Ensemble
volatility $\sigma_i^{\mathrm{ens}}$ is the mean, over the $K$ paths, of
the standard deviation of log-returns along each path. The prior gap is
$|\sigma_i^{\mathrm{ens}}-\sigma_i^{\mathrm{id}}|/\sigma_i^{\mathrm{id}}$.
A smaller value means the samples sit closer to that product's typical
scale. History gap compares the ensemble with the recent window. Let
$r_i^{\mathrm{hist}}$ be the log-return over the last horizon-length
stretch of the context, and let $r_i^{\mathrm{ens}}$ be the mean
terminal log-return across the $K$ paths. The history gap is
$|r_i^{\mathrm{ens}}-r_i^{\mathrm{hist}}|/s_i$, where $s_i$ is the same
product's historical return scale $\sigma_i^{\mathrm{id}}$ times the
square root of the horizon length. A smaller value means the samples sit
closer to the recent path.

Spread is the diversity of the ensemble: the standard deviation of the
$K$ terminal log-returns, divided by the same $s_i$. A smaller value
means a tighter set of paths.
Table~\ref{tab:app-cfg-sweep} records the sweep.

\begin{table*}[h]
\centering
\caption{\textbf{Guidance sweep averaged over 100 products.}}
\label{tab:app-cfg-sweep}
\small
\setlength{\tabcolsep}{5pt}
\begin{tabular}{r rrr rrr}
\toprule
& \multicolumn{3}{c}{Identity $w_{\mathrm{id}}$ ($w_{\mathrm{hist}}{=}1$)}
& \multicolumn{3}{c}{History $w_{\mathrm{hist}}$ ($w_{\mathrm{id}}{=}1$)} \\
\cmidrule(lr){2-4}\cmidrule(lr){5-7}
$w$ & Prior $\downarrow$ & Hist. $\downarrow$ & Spread $\downarrow$
    & Prior $\downarrow$ & Hist. $\downarrow$ & Spread $\downarrow$ \\
\midrule
1 & 0.41 & 0.52 & 1.08 & 0.41 & 0.52 & 1.08 \\
2 & 0.28 & 0.49 & 0.91 & 0.39 & 0.37 & 0.93 \\
3 & 0.20 & 0.47 & 0.78 & 0.37 & 0.26 & 0.81 \\
4 & 0.15 & 0.46 & 0.69 & 0.36 & 0.19 & 0.72 \\
\bottomrule
\end{tabular}
\end{table*}

The two weights move different margins of the same ensemble.
Raising $w_{\mathrm{id}}$ from 1 to 4 cuts the prior gap from 0.41 to
0.15, so the samples move toward each product's own volatility, while
the history gap stays near 0.5.
Raising $w_{\mathrm{hist}}$ from 1 to 4 cuts the history gap from 0.52
to 0.19, so the samples move toward the recent window, while the prior
gap stays near 0.4.
On both axes the spread falls, from 1.08 at $w{=}1$ to 0.69 under
identity guidance and to 0.72 under history guidance: fidelity to the
selected context comes with a less diverse ensemble. We additionally compared the IC and RankIC for return forecasting. A larger CFG allows predictions to be more concentrated for certain products, and we also observed that for some high-volatility products, increasing CFG improves forecasting performance. However, the average IC scores remain largely unchanged overall. The main results therefore keep both weights at 1.

\section{More Ablation Study}
\label{app:ablation}
We further conduct ablation experiments to validate the design choices of each component in our KiT: the candlestick representation, whether the window-constant identity signals (market, sector, instrument and timescale) are injected through the AdaLN trunk, and whether the per-bar signals (clock, session, day of week, month, day of year and event flags) are injected into the token embeddings. Each variant keeps the data protocol and training recipe of the main model fixed and changes a single factor, and all are scored by the mean return RankIC.

\begin{table}[htbp]
\centering
\caption{\textbf{Ablation on the candlestick representation.} Each row alters
only the input state; all other settings follow the main model.}
\label{tab:app-ablate-repr}
\small
\renewcommand{\arraystretch}{1.15}
\setlength{\tabcolsep}{6pt}
\begin{tabular}{lc}
\toprule
Representation of a bar & Mean return RankIC $\uparrow$ \\
\midrule
\textbf{Five-dimensional log-ratio state (\method{})} & \textbf{0.0571} \\
\midrule
Close-only log-price input & 0.0183 \\
Naive OHLCV input (normalized) & 0.0209 \\
\bottomrule
\end{tabular}
\end{table}

Table~\ref{tab:app-ablate-repr} validates the candlestick representation.
Encoding a bar as raw OHLCV levels, even after robust normalisation, discards the
scale-free relative form of the candle and lowers the mean RankIC to $0.0209$;
reducing the input further to a close-only log-price, which also drops the
intrabar range and volume, gives the weakest score of $0.0183$.

\begin{table}[htbp]
\centering
\caption{\textbf{Ablation on conditioning injection.} Identity denotes the
window-constant signals (market, sector, instrument, timescale); Calendar
denotes the per-bar signals (clock, session, day of week, month, day of year,
event flags).}
\label{tab:app-ablate-cond}
\small
\renewcommand{\arraystretch}{1.1}
\setlength{\tabcolsep}{6pt}
\begin{tabular}{lc}
\toprule
Configuration & Mean return RankIC $\uparrow$ \\
\midrule
\textbf{AdaLN identity + per-bar calendar (\method{})} & \textbf{0.0571} \\
\midrule
\emph{Identity $\rightarrow$ AdaLN} & \\
\quad Market and sector only (drop instrument, scale) & 0.0386 \\
\quad w/o identity & 0.0329 \\
\midrule
\emph{Per-bar condition $\rightarrow$ token embeddings} & \\
\quad w/o per-bar condition & 0.0372 \\
\bottomrule
\end{tabular}
\end{table}

Table~\ref{tab:app-ablate-cond} shows that both conditioning pathways contribute.
Withdrawing the identity signals entirely ($\vc_{\mathrm{base}}{=}\vzero$) costs
the most, dropping the mean RankIC from $0.0571$ to $0.0329$, while keeping only
the coarse market and sector identifiers recovers to $0.0386$: the fine-grained
instrument and timescale embeddings therefore carry signal that the coarse prior
cannot. Removing the per-bar calendar signals similarly degrades the score to
$0.0372$, indicating that clock, session and event context supply temporal
structure that neither identity conditioning nor the observed history provides on
its own. Together these results justify the full design, in which each class of
signal is injected at the granularity that matches its nature.
\end{document}